\pdfoutput=1

\documentclass[11pt]{article}

\usepackage[]{emnlp2021}

\usepackage{times}
\usepackage{latexsym}

\usepackage[T1]{fontenc}

\usepackage[utf8]{inputenc}

\usepackage{microtype}

\usepackage{graphicx}
\usepackage{amsmath}
\usepackage{makecell}
\usepackage{booktabs}
\usepackage{amssymb}
\usepackage{multirow}

\title{Improving Dialogue State Tracking by Joint Slot Modeling}

\author{
Ting-Rui Chiang \and Yi-Ting Yeh \\
Carnegie Mellon University \\ 
\texttt{\{tingruic,yitingye\}@andrew.cmu.edu}
}

\begin{document}
\maketitle
\begin{abstract}

Dialogue state tracking models play an important role in a task-oriented dialogue system.
However, most of them model the slot types conditionally independently given the input.
We discover that it may cause the model to be confused by slot types that share the same data type.
To mitigate this issue, we propose TripPy-MRF and TripPy-LSTM that models the slots jointly.
Our results show that they are able to alleviate the confusion mentioned above, and they push the state-of-the-art on dataset MultiWoZ 2.1 from 58.7 to 61.3.
Our implementation is available at \url{https://github.com/CTinRay/Trippy-Joint}. 

\end{abstract}

\section{Introduction} \label{sec:intro}
Multi-domain dialogue state tracking (DST) has been an important challenge of conversational artificial intelligence~\cite{ram2018conversational}.
Compared to single-domain DST, the belief state of a multi-domain DST contains slot types across multiple domains.
It causes difficulty to the traditional DST models that assume the complete ontology is available because a complete ontology becomes hard to obtain~\cite{wu2019global} when there are many domains,

To tackle these challenges, many of the mainstream approaches for DST formulate this as a span prediction task~\cite{xu-hu-2018-end, wu-etal-2019-transferable, kumar2020ma}.
Instead of predicting a predefined value in the ontology, the span-based DST models predict slot values by extracting a span from the conversation between the user and the system.
By directly extracting spans as slot values, the DST models are able to handle unseen slot values and are potentially transferable to different domains.
However, the span-based DST models can only deal with the slot values that are explicitly expressed as a sequence of tokens and fall short in dealing with coreference ("I'd like a restaurant in the same area") and implicit choice ("Any of those is ok")
The current state-of-the-art DST model TripPy~\cite{heck-etal-2020-TripPy} is designed to deal with these problems.
In addition to extracting values directly from the user utterance, TripPy maintains two additional memories on the fly and uses them to deal with the coreference and implicit choice challenges.
The design details of TripPy will be further illustrated in Section~\ref{subsec:TripPy}

Despite the success of these span-based models, one concern is that they often consider the slots conditionally independently.
For example, in TripPy, the conversation is first encoded by BERT~\cite{devlin-etal-2019-bert}, and then the extracted features are used by n modules for the n slot types independently.
That implies, conditioning the features extracted by BERT, the slots are predicted independently.
This may increase the difficulty of this task because these slot types are indeed not independent of each other.
If the value of one slot type in a domain is present in an utterance, then it is more likely that the value of another slot type of the same domain is present.

In particular, in this work, we focus on the cases where modeling the joint probability of the slots could possibly be helpful.
We notice that there are some slot types that share the same data type.
The data types of both the slot type "hotel-people" and the slot type "restaurant-book people" are both integers.
Therefore, when an integer is present in an utterance, a DST model needs to predict which slot(s) this integer is for.
Given the fact that some slots co-occur more often than others, modeling the slot types jointly may be helpful when making this kind of prediction.
Based on this intuition, we group the slot types into 4 groups by their data type: time, place name, integer.
We then analyze the current state-of-the-art model TripPy~\cite{heck-etal-2020-TripPy,MehriDialoGLUE2020}.
Our analysis shows that TripPy is indeed often confused with slot types in the same group.

We then investigate whether modeling the relation between slot types explicitly can alleviate this problem.
By modeling the slot types jointly, information from the slots that the model is more confident with may reduce the confusion for the other slots.
Specifically, to achieve this goal, we propose to model the joint probability of slots with MRF and LSTM models.
They can be plugged into any DST model that models the slots conditionally independently.
In this work, we choose the current state-of-the-art DST model, TripPy pre-trained with DialoGLUE \cite{MehriDialoGLUE2020}.
We experiment on the most widely used multi-domain DST dataset MultiWoZ 2.1 \cite{zang2020MultiWoZ,eric2019MultiWoZ}.
The results show that both the MRF and the LSTM models are able to mitigate the confusion, and can improve the performance consistently.
We push the state-of-the-art performance on MultiWoZ 2.1 from 58.7 to 61.3.

\section{Background}

\subsection{Problem Formulation}
The dialogue state tracking (DST) task is to predict values of slot types for each turn in task-oriented dialogue. Specifically, dialogue consists of a sequence of user and system utterances.
The input of the task is a dialogue $X = \{(U_1, M_1), \cdots, (U_T, M_T)\}$,
where $U_t$ and $M_t$ are utterances by user and system respectively, and $T$ is the total number of turns in this dialogue.
The output of the task is the dialogue state at each time step.
A dialogues state consists of $N$ slot types.
The goal of a DST task is to predict the slot values of those $N$ slot types at each time step.
The slot value $S^{(t)}_{s}$ of the s-th slot type at time step $t$ might be a span in the previous conversation of simply \textit{none}.
We denote the state at time step $t$ as $S^{(t)} = \{S^{(t)}_{1}, S^{(t)}_2, \cdots, S^{(t)}_N \}$.

\subsection{TripPy: Triple Copy Strategy for DST} \label{subsec:TripPy}

As described in Section~\ref{sec:intro}, in addition to extract values from the user utterance, TripPy maintains two memories to tackle the coreference and the implicit choice problems in the span-based DST model.
Specifically, we use \textit{system inform memory} to remember the slots which the model previously informed, and \textit{DS memory} to store the seen slots in the dialogue history.
At dialogue turn $t$, TripPy predicts the belief state $S^{(t)}$ with the following steps:
1) The model first uses a pretrained BERT model to encode the dialogue history until the current time step $t$.
2) For each slot type $s$, it uses a classifier to predict the \textit{slot class} $C^{(t)}_s$.
For slot types whose possible values are "yes" or "no", possible values of $C^{(t)}_s$ are \{\textit{none}, \textit{dontcare}, \textit{yes}, \textit{no}\}.
For the other slot types, possible values of $C^{(t)}_s$ are \{\textit{none}, \textit{dontcare}, \textit{span}, \textit{inform}, \textit{refer}\}. 
These slot classes determine which source the slot value should be copied from.
3) The first two slot classes express special cases.
\textit{none} denotes that the slot does not take a value in this turn, and \textit{dontcare} states that any value is acceptable for this slot.
If $C^{(t)}_s$ is \textit{span}, then TripPy uses a span prediction model to copy a span of text from the dialogue history as the prediction of a slot type.
On the other hands, if $C^{t}_s$ is \textit{inform} and \textit{refer}, the model will instead copy the value from \textit{system inform memory} and \textit{DS memory} respectively.
The \textit{system inform memory} allows the model to solve the implicit choice issue and the \textit{DS memory} helps the model solve coreference problems.

\subsection{Experimental Setup}
We train and test our model on MultiWoZ 2.1 \cite{eric2019MultiWoZ}, which is the most challenging dataset for DST and is widely used for the evaluation of multi-domain DST models.
MultiWoZ 2.1 is comprised of over 10000 dialogues in 5 domains and has 30 different slots with over 45000 possible values.
We compute the joint goal accuracy (JGA) on all test samples to evaluate our models.
The JGA is the ratio of dialog turns in the dataset for which all slots have been filled with the correct value according to the ground truth.
Our baseline is the TripPy model in \cite{MehriDialoGLUE2020}, which is additionally pretrained with more dialogue tasks.
To reproduce their result, we follow the setting of their hyperparameters and successfully achieve JGA 58.0, which is close to the reported result of 58.7 in the paper.

\section{Analysis}

\subsection{Accuracy per Copy Class}

\begin{table*}[]
\tiny
    \centering
    \begin{tabular}{l|ccccccc|cc}
    \toprule
    Slot Type & none & don't care & span & inform & refer & true & false & Oracle & $\Delta$ \\
        \midrule
taxi-leaveAt            & .99 (7218) &  -  (0)  & .88 (131) & .91 (23) & .00 (2) & . -  (0) & . -  (0)   & .581 & .001  \\
taxi-destination        & .98 (7159) &  -  (0)  & .98 (108) & .94 (18) & .61 (89) & . -  (0) & . -  (0)    & .587 & .007\\
taxi-departure          & .98 (7169) &  -  (0)  & .89 (100) & 1.0 (21) & .60 (84) & . -  (0) & . -  (0)    & .587 & .007\\
taxi-arriveBy           & .99 (7278) & .50 (2)  & .85 (55) & 1.0 (10) & .76 (29) & . -  (0) & . -  (0)    & .581 & .001\\
restaurant-book\_people & .99 (7001) &  -  (0)  & .99 (319) & 1.0 (41) & .85 (13) & . -  (0) & . -  (0)    & .583 & .002\\
restaurant-book\_day    & 1.0 (7011) &  -  (0)  & .99 (286) & 1.0 (53) & .92 (24) & . -  (0) & . -  (0)    & .581 & .001\\
restaurant-book\_time   & .99 (6979) &  -  (0)  & .98 (328) & 1.0 (67) & . -  (0) & . -  (0) & . -  (0)      & .583 & .003\\
restaurant-food         & .99 (6801) & .45 (20) & 1.0 (422) & .98 (131) & . -  (0) & . -  (0) & . -  (0)   & .593 & .013\\
restaurant-pricerange   & .99 (6906) & .78 (9)  & .98 (343) & .98 (113) & 1.0 (3) & . -  (0) & . -  (0)   & .587 & .006\\
restaurant-name         & .96 (6926) & .00 (2)  & .93 (162) & 1.0 (283) & .00 (1) & . -  (0) & . -  (0)   & .598 & .018\\
restaurant-area         & .98 (6881) & .71 (17) & .97 (314) & .96 (137) & .88 (25) & . -  (0) & . -  (0) & .594 & .014\\
hotel-book\_people      & .99 (7031) &  -  (0)  & .99 (300) & .97 (35) & .62 (8) & . -  (0) & . -  (0)     & .581 & .001\\
hotel-book\_day         & 1.0 (7035) &  -  (0)  & .98 (289) & 1.0 (40) & .70 (10) & . -  (0) & . -  (0)    & .582 & .002\\
hotel-book\_stay        & .99 (7015) &  -  (0)  & .99 (314) & 1.0 (45) & . -  (0) & . -  (0) & . -  (0)      & .582 & .001\\
hotel-name              & .96 (6884) & .00 (1)  & .98 (170) & 1.0 (319) & . -  (0) & . -  (0) & . -  (0)    & .59  & .009\\
hotel-area              & .97 (7028) & .45 (29) & 1.0 (205) & .98 (92) & .65 (20) & . -  (0) & . -  (0)  & .595 & .014\\
hotel-parking           & .97 (7204) & .33 (6)  & . -  (0) & . -  (0) & . -  (0) & .86 (154) & .50 (10)     & .587 & .007\\
hotel-pricerange        & .98 (6986) & .67 (21) & .97 (271) & 1.0 (91) & .60 (5) & . -  (0) & . -  (0)   & .59  & .01 \\
hotel-stars             & .99 (7094) & .50 (6)  & .98 (210) & .91 (64) & . -  (0) & . -  (0) & . -  (0)     & .584 & .004\\
hotel-internet          & .98 (7200) & 1.0 (2)  & . -  (0) & . -  (0) & . -  (0) & .89 (169) & .67 (3)      & .588 & .007\\
hotel-type              & .93 (7135) & .00 (5)  & . -  (0) & . -  (0) & . -  (0) & .59 (80) & .85 (154)     & .612 & .031\\
attraction-type         & .98 (6835) & .38 (8)  & .98 (430) & 1.0 (101) & . -  (0) & . -  (0) & . -  (0)    & .595 & .015\\
attraction-name         & .98 (7006) & .00 (2)  & .97 (137) & 1.0 (229) & . -  (0) & . -  (0) & . -  (0)    & .59  & .01 \\
attraction-area         & .98 (6946) & .39 (18) & .98 (248) & .96 (135) & .93 (27) & . -  (0) & . -  (0) & .593 & .013\\
train-book\_people      & .98 (7040) &  -  (0)  & .96 (314) & 1.0 (8) & .42 (12) & . -  (0) & . -  (0)     & .591 & .011\\
train-leaveAt           & .97 (7125) & .50 (2)  & .96 (228) & .68 (19) & . -  (0) & . -  (0) & . -  (0)     & .612 & .032\\
train-destination       & .98 (6693) &  -  (0)  & .99 (503) & 1.0 (176) & .00 (2) & . -  (0) & . -  (0)    & .586 & .006\\
train-day               & .99 (6791) &  -  (0)  & 1.0 (481) & 1.0 (86) & .94 (16) & . -  (0) & . -  (0)    & .585 & .005\\
train-arriveBy          & .99 (7034) & .75 (4)  & .98 (322) & .75 (12) & .00 (2) & . -  (0) & . -  (0)    & .592 & .012\\
train-departure         & .98 (6749) &  -  (0)  & .99 (478) & 1.0 (146) & .00 (1) & . -  (0) & . -  (0)    & .587 & .007\\
    \midrule
    sum                & 3928 & 77 & 168 & 73 & 29 & 116 & 37 \\
    \bottomrule
    \end{tabular}
    \caption{Accuracy of each slot type by ground truth class. The number in the parenthesis is the number of occurrence of the class for the slot type. The oracle column is the performance calculated when presuming the slot value is predicted correctly.}
    \label{tab:slot-accuracy}
\end{table*}

\begin{table}[]
    \centering
    \begin{tabular}{l|ccc}
    \toprule
     & Other & FN & FP \\
    \midrule
taxi-leaveAt & 13 & 7 & 25           \\
taxi-destination & 10 & 32 & 13      \\
taxi-departure & 14 & 37 & 13        \\
taxi-arriveBy & 1 & 15 & 19          \\
restaurant-book\_people & 1 & 5 & 37 \\
restaurant-book\_day & 1 & 3 & 16    \\
restaurant-book\_time & 3 & 4 & 12   \\
restaurant-food & 3 & 13 & 48        \\
restaurant-pricerange & 4 & 8 & 57   \\
restaurant-name & 9 & 7 & 45         \\
restaurant-area & 3 & 19 & 57        \\
hotel-book\_people & 2 & 6 & 34      \\
hotel-book\_day & 1 & 7 & 8          \\
hotel-book\_stay & 2 & 2 & 17        \\
hotel-name & 6 & 3 & 72              \\
hotel-area & 2 & 24 & 94             \\
hotel-parking & 11 & 20 & 88         \\
hotel-pricerange & 5 & 13 & 38       \\
hotel-stars & 4 & 10 & 24            \\
hotel-internet & 1 & 18 & 85         \\
hotel-type & 12 & 49 & 220           \\
attraction-type & 2 & 11 & 104       \\
attraction-name & 21 & 5 & 31        \\
attraction-area & 0 & 25 & 53        \\
train-book\_people & 1 & 19 & 37     \\
train-leaveAt & 3 & 14 & 57          \\
train-destination & 6 & 4 & 12       \\
train-day & 1 & 2 & 17               \\
train-arriveBy & 3 & 9 & 24          \\
train-departure & 3 & 6 & 8          \\
\midrule
 & 148 & 397 & 1365                  \\
    \bottomrule
    \end{tabular}
    \caption{Number of incorrect slot value predictions in different cases of \textit{none} prediction. FN (false negative)/FP (false positive) means that the ground truth class is \textit{none}/is not \textit{none}, but the model predicts the class incorrectly.}
    \label{tab:none-accuracy}
\end{table}

We analyze the accuracy of each slot by the ground truth copy strategy to use on the validation set.
For each slot type, we also calculate the joint goal accuracy presuming it is predicted correctly.
From the results in Table~\ref{tab:slot-accuracy} we can have a few observations.
1) The class \textit{none} takes most of the cases. Therefore, it is important to predict whether the value is \textit{none} correctly.
2) On the other hand, improving other classes may be less effective, since the number of mistakes is much smaller.

The above results encourage us to inspect the prediction of the class \textit{none} more closely.
Here we focus on the cases where the model is to decide whether it is going to give a value to a slot type. 
We ignore cases where the value of the slot type in the previous turn is not none.
We calculate the number of incorrect value predictions when \textit{none} is predicted incorrectly.
From Table~\ref{tab:none-accuracy}, we can see that many incorrect predictions are resulted from incorrect \textit{none} prediction.

\subsection{Relation Between Slots} \label{subsec:incorrect_freq}

\begin{figure*}
    \centering
    \includegraphics[width=0.70\linewidth]{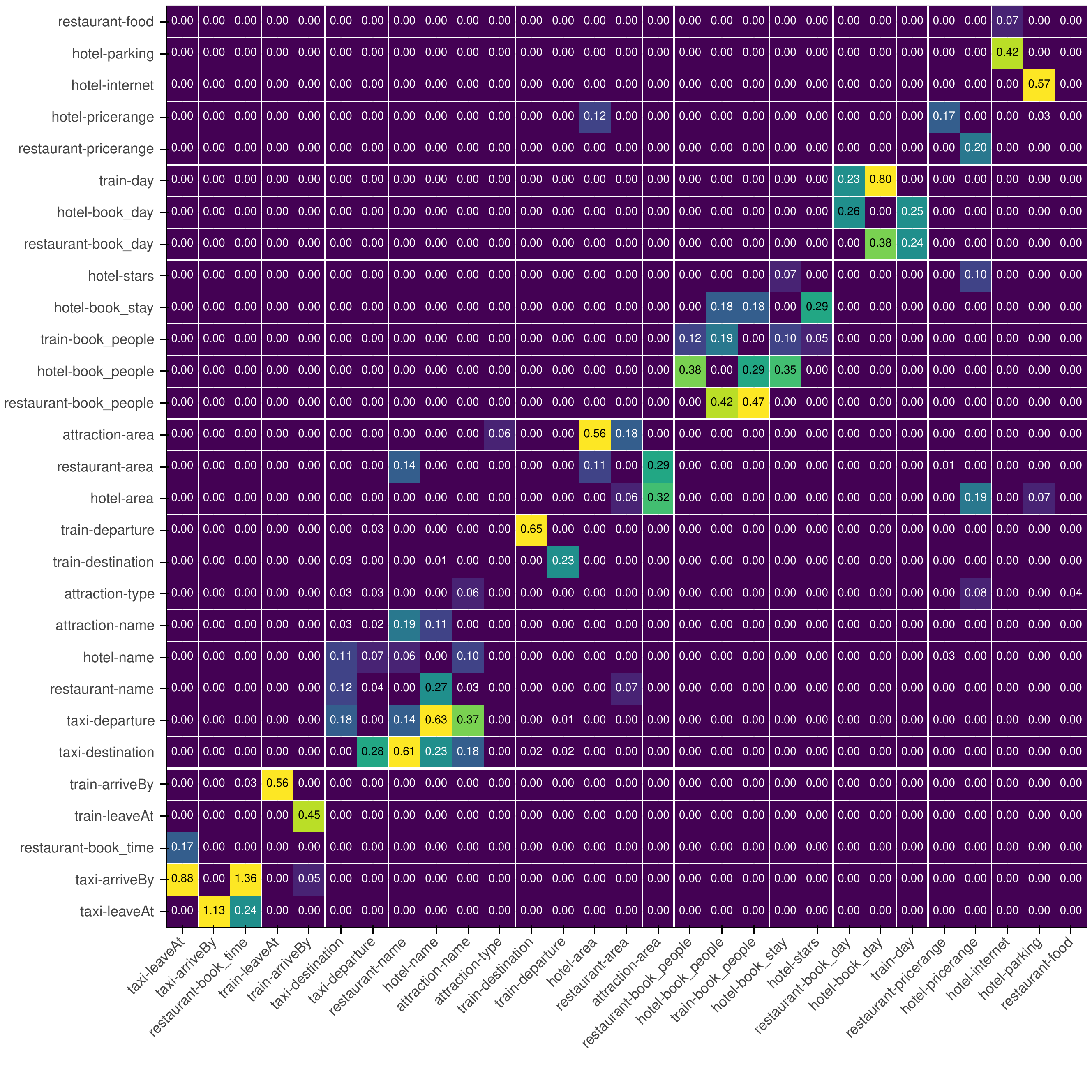}
    \caption{The value of $p_1 + p_2$ as described in section \ref{subsec:incorrect_freq}.}
    \label{fig:violation}
\end{figure*}

We further examine the model's behavior when \textit{none} is not predicted correctly.
We check the frequency of the following cases of incorrect prediction:
1) $p_1$: when the ground truth class is \textit{none}, the cases where the predicted slot value is the ground truth value of another slot type.
2) $p_2$: when the predicted class is \textit{none}, cases of the slot value that is the ground truth value of another slot type.
We plot the frequency $p_1 + p_2$ in Figure~\ref{fig:violation}.
From the figure, we can see that the model is often confused by similar slot types.
For example, the value of the slot type "taxi-departure" and the value of the slot type "taxi-ariveBy" is time.
It may be the reason why the model often fails to give non-\textit{none} value to the correct slot type.

\begin{figure*}[h!]
    \centering
    \includegraphics[width=0.5\linewidth]{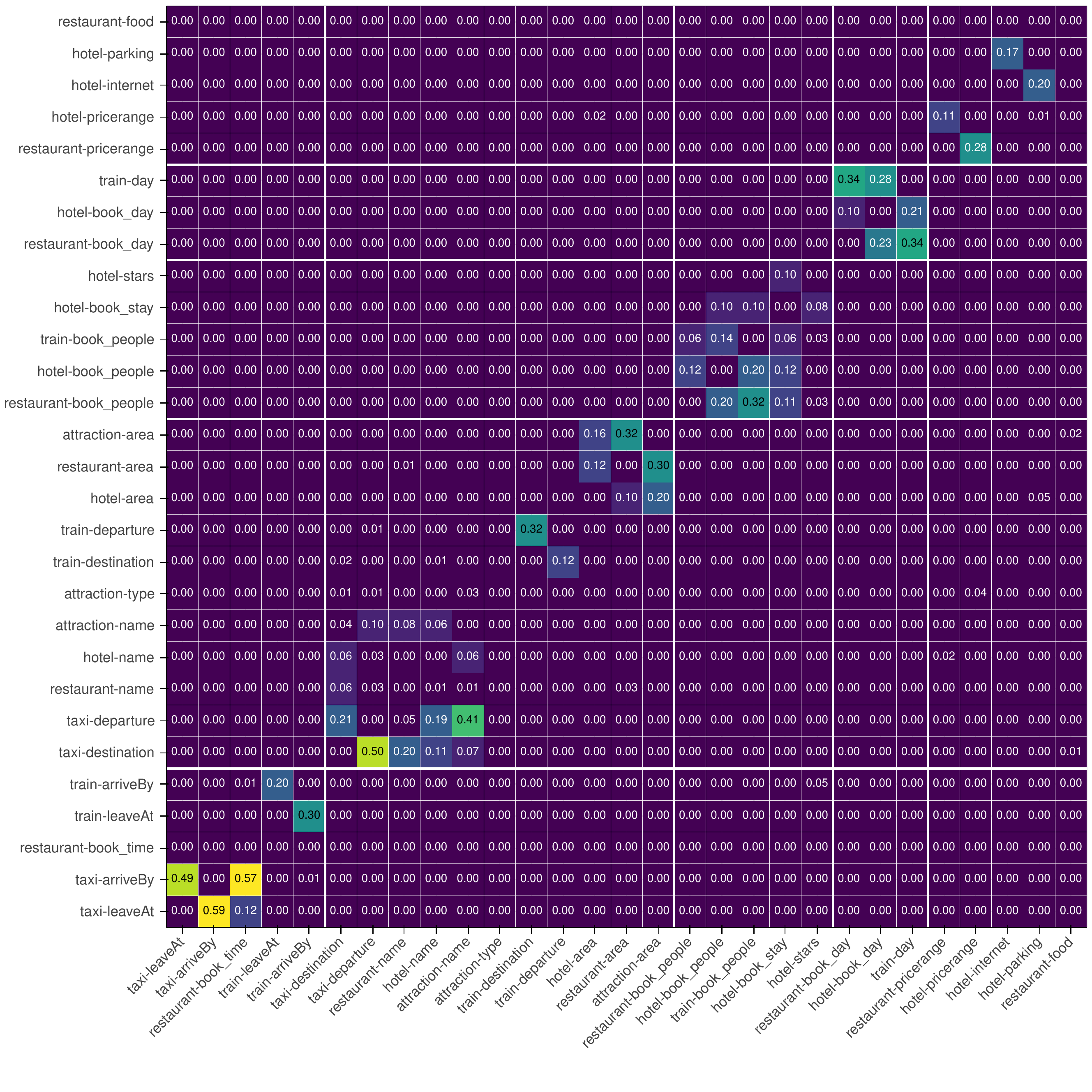}
    \caption{The value of $p_1 + p_2$ (Section~\ref{subsec:incorrect_freq}) predicted by the TripPy + MRF model.}
    \label{fig:mrf_violation}
\end{figure*}

\begin{figure*}[h!]
    \centering
    \includegraphics[width=0.5\linewidth]{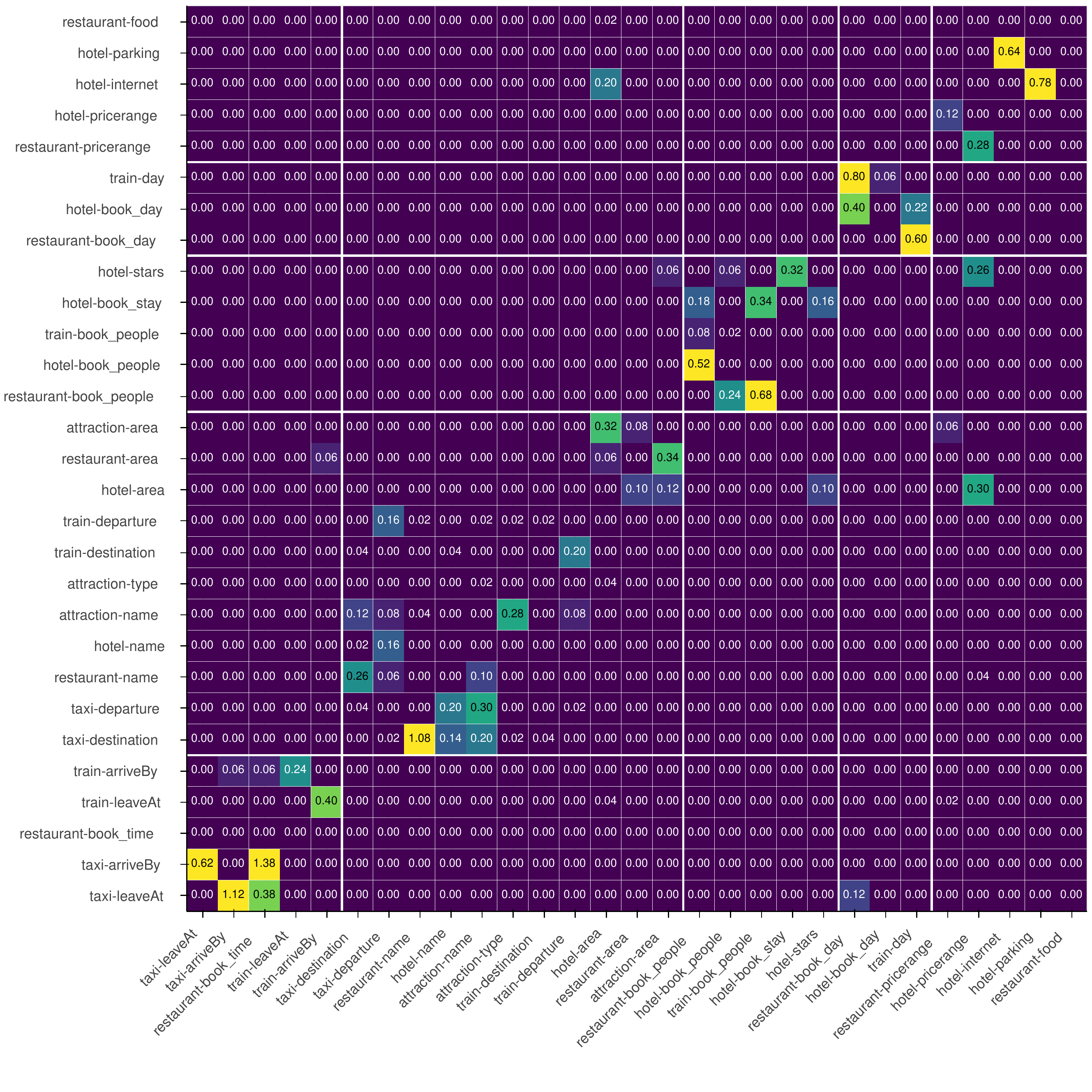}
    \caption{The value of $p_1 + p_2$ (Section~\ref{subsec:incorrect_freq}) predicted by the TripPy + LSTM model.}
    \label{fig:lstm_violation}
\end{figure*}

\section{Proposed Method}

We conjecture some mistakes made by TripPy maybe because it models the slots conditionally independently.
Based on Table~\ref{tab:none-accuracy}, the model does not predict \textit{none} very accurately, and Figure~\ref{fig:violation} indicates that it is because the model is often confused by similar slot types.
In the original design, given a dialogue $X$, TripPy models the class prediction independently as follow:
\begin{equation}
    P(C^{(t)}_1, C^{(t)}_2, \cdots, C^{(t)}_N |X) = \prod_{s = 1}^{N} P(C^{(t)}_s | X).
\end{equation}
TripPy ignores the potential correlation between different slots.
Thus, for all $s$, $P(C^{(t)}_s | X)$ needs to predict the class correctly at the same time.
It may be difficult since there are numerous slots.

To alleviate this problem, we propose to directly model the joint probability $P(C^{(t)}_1, C^{(t)}_2, \cdots, C^{(t)}_N |X)$.
If $c_1, c_2, \cdots, c_N$ are the correct classes, modeling the joint probability does not require
$P(C^{(t)}_s = c_s) > P(C^{(t)}_s = c'_s)$ for all $s$ and all incorrect assignments $\{c'_s\}_{s=1}^N$.
Instead, to predict the belief state correctly, the model is only required to ensure
$P(C^{(t)}_1=c_1, C^{(t)}_2=c_2, \cdots, C^{(t)}_N=c_N |X) > P(C^{(t)}_1 = c'_1, C^{(t)}_2 = c'_2, \cdots, C^{(t)}_N = c'_N |X)$.
In this way, information from the other states could possibly help the prediction of difficult slot types.
For example, if the model has high confidence that the utterance contains the value of slot "restaurant-name", the number in the utterance is more likely the value of slot "restaurant-book people".
Therefore, confusion shown in Figure~\ref{fig:violation} could possibly be alleviated.

\subsection{Modeling the Joint Probability with Graphical Models}

One approach that models the joint probability is to use a Markov random field (MRF).
This approach is parameter-efficient and incurs little computation overhead for inference.
However, having a large number of vertices in an MRF makes training intractable.
Therefore, we follow the intuition above, grouping the slots by their domains, and model the joint distribution within each group respectively.
Since a large portion of the mistakes made by the original TripPy model is due to the incorrect prediction of whether a slot value is present in an utterance, we further decompose the probability of a class as
\begin{equation}
\begin{split}
    &P(C^{(t)}_s = c) \\
    &=  P(C^{(t)}_s = c | C^{(t)}_s \ne \text{"\textit{none}"}) P(C^{(t)}_s \ne \text{"\textit{none}"} )
\end{split}
\end{equation}
Let $C'^{(t)}_s$ be the random variable indicating whether $C^{(t)}_s = \text{"\textit{none}"}$.
Modeling only the $P(\{ C'^{(t)}_s \}_{s=1}^{N})$ part jointly makes the training and inference of the MRF model feasible: given input $x$, for each domain $d$ and the assignment of slots $c_1, c_2, \cdots, c_N$ for the $N$ slots in $d$, the probability of the assignment is
\begin{align}
    \begin{split}
        & P(c_1, c_2, \cdots, c_N | x) \\ 
        &= \frac{1}{Z(x)} \Phi_{d}(\{C'^{(t)}_s = c'_s\}_{s=1}^N) \\
        & \prod_{s=1}^N \Phi_s(c_s, x) P(C^{(t)}_s = c_s | C'^{(t)}_s = c'_s),
    \end{split}
    \label{eq:potential}
\end{align}
where $\Phi_{d}$ map each assignment $c'_1, c'_2, \cdots, c'_N$ to a learnable scalar, and $\Phi_s(c'_s, x)$ is a scalar resulted from applying a linear layer over the pooled BERT output.
Since the number of slot types in a domain is at most 10, and all of the variables are binary, the partition function $Z_i(x)$ can be calculated by enumerating all the possible assignments of $c'_1, c'_2, \cdots, c'_N$.
Therefore, we can train this model with the plain MLE loss.

\subsection{Modeling with LSTM models}

Another approach we explore is to use the LSTM model~\cite{schmidhuber1997long}.
Similar to language modeling, we model the joint probability by predicting the value of each slot sequentially:
\begin{equation}
    \begin{split}
    &P(C^{(t)}_1, C^{(t)}_2, \cdots, C^{(t)}_N |X) \\ \
   &= \prod_{s = 1}^{N} P( C^{(t)}_s | X, C^{t}_1, \dots, C^{t}_{s-1}),
   \end{split}
   \label{eq:lstm}
\end{equation}
For the implementation of this model, we first encode the dialogue history $X$ with BERT and extract its CLS token representation $h_X$.
The original TripPy model will feed $h_X$ into a linear layer $f_{s}$ to predict slot class for each slot.
\begin{equation}
    P(C^{(t)}_s|X) = Softmax(f_{s}(h_X))
\end{equation}
Here, for all $s$, we additionally train another linear layer $f^{lstm}_{s}$ to transform $h_X$ and feed it into a LSTM layer:
\begin{equation} \label{eq:lstm}
    h_s = LSTM(f^{lstm}_{s}(h_X), h_{s-1})
\end{equation}
We then concatenate $h_s$ and $h_x$ to model the conditional probability:
\begin{equation} \label{eq:predict_lstm}
    P(C^{(t)}_s | X, C^{(t)}_1, \dots, C^{(t)}_{s-1}) = f_{s}([h_X;h_s])
\end{equation}
where $[x;y]$ denotes the concatenation of vectors $x$ and $y$.
We name this model TripPy+LSTM.


\section{Results and Discussion}

\begin{table}[]
    \centering
    \begin{tabular}{l c c}
    \toprule
        Model & Dev. & Test \\
    \midrule
        TripPy~\cite{MehriDialoGLUE2020} & & 58.7 \\
        TripPy (reproduced) & 61.3 & 58.0 \\
         + MRF & 59.3 & 60.1 \\
         + LSTM & 62.8  & 61.3 \\
    \bottomrule
    \end{tabular}
    \caption{Joint goal accuracy on MultiWoZ 2.1.}
    \label{tab:main_result}
\end{table}


Our main results are shown in Table~\ref{tab:main_result}.
Both MRF and LSTM modules have a consistent improvement over the baseline model on the test set of MultiWoZ 2.1, which proves the effectiveness of our proposed approaches.

As a more fine-grained analysis, we plot the frequency of incorrect prediction as described in Section~\ref{subsec:incorrect_freq}.
Compared to Figure~\ref{fig:violation}, Figure~\ref{fig:mrf_violation} and Figure~\ref{fig:lstm_violation} show that both TripPy+MRF and TripPy+LSTM can relieve the confusion between slot types that have the same data type.
This is coherent to our intuition for modeling the slot types jointly.

The results also show that TripPy+LSTM performs better than TripPy+MRF.
It may be because TripPy+LSTM models all the assignment $\{ C_s \}_{s=1}^N$ of the classes jointly, while TripPy+MRF only models $\{ C'_s \}_{s=1}^N$ jointly.
Also, TripPy+LSTM has the potential to model the joint distribution across the domains.
However, this also sacrifices efficiency.
Since TripPy+LSTM predicts the slots sequentially, it cannot predict the slots in parallel.
TripPy+MRF could be a better choice when latency is an important issue.

 
\section{Related Work}
\label{sec:related_work}

\citet{williams-etal-2013-dialog} is the first benchmark for DST models.
After that, \citet{henderson-etal-2014-second} proposed the DSTC2 dataset which includes more linguistic phenomena.
\citet{budzianowski-etal-2018-multiwoz} proposed MultiWoZ, which is the first large-scale human-human multi-domain task-oriented DST dataset.
Its successive versions \cite{eric2019MultiWoZ,zang-etal-2020-MultiWoZ} cleanse the noisy conversations in it.
Since then, it has become the standard DST benchmark dataset.

Early approaches for DST rely on a list candidate slot values \cite{mrksic-etal-2017-neural,Liu2017,ramadan-etal-2018-large,rastogi2017scalable}.
Some recent approaches formulate it as a sequence generation task.
TRADE \cite{wu-etal-2019-transferable} and MA-DST \cite{kumar2020ma} generates slot values with a pointer generator with a copy mechanism.
SOM-DST \cite{kim-etal-2020-efficient} uses a BERT model to encode the dialogue. To improve the efficiency, the state from the previous turn instead of the whole dialogue history is fed into the model.
SUMBT \cite{lee-etal-2019-sumbt} utilizes a BERT model \cite{devlin-etal-2019-bert} to encode both the dialogue history and candidate slot values.
In \citet{gao-etal-2019-dialog} they formulate DST as a question answering task.
DS-DST \cite{zhang-etal-2020-find} adopts a dual strategy: they predict slot values from candidates but also predict slot values by copying from the dialogue.
TripPy \cite{heck-etal-2020-TripPy} predicts slot values by copying text span from three different sources.
In \citet{MehriDialoGLUE2020}, they improve the performance of TripPy by pre-training the BERT with multiple dialogue tasks.
Simple-TOD \cite{hosseini2020simple} and SOLOIST \cite{peng2020soloist} utilize GPT-2 for end-to-end slot value and response generation.
In \citet{SHIYANG2020CoCoCC}, they investigate the generalization capability of DST models for unseen slot values.
MinTL \cite{lin2020mintl} augments data by substituting slot values in the utterances and modifying the syntax structure.

\section{Conclusion}
In this work, we explore the relation between slot types.
Our contributions are two-fold.
1) We identify a problem with the current state-of-the-art model TripPy.
We observe that it is often confused by slots that share the same type of value.
2) We propose TripPy+MRF and TripPy+LSTM that aim at mitigating this problem.
One can choose one of them based on the trade-off between efficiency and accuracy.
Our preliminary results show that both of them are able to alleviate the confusion.
Our findings in this work lead future researchers in a promising direction to improve the performance of multi-domain DST.

\section*{Acknowledgments}
We would like to thank Graham Neubig for his in-depth discussions. We are also thankful to the anonymous reviewers for their comments on the paper.


\bibliography{anthology, custom}
\bibliographystyle{acl_natbib}

\appendix

\end{document}